\documentclass[runningheads]{llncs}
\usepackage{subfig}
\usepackage{graphicx}
\usepackage{hyperref}

\usepackage{tikz}
\usepackage{comment}
\usepackage{amsmath,amssymb}
\usepackage{color}

\usepackage[accsupp]{axessibility}

\usepackage[width=122mm,left=12mm,paperwidth=146mm,height=193mm,top=12mm,paperheight=217mm]{geometry}

\begin{document}
\pagestyle{headings}
\mainmatter

\title{One-Shot Face Reenactment on Megapixels}

\titlerunning{One-Shot Face Reenactment on Megapixels}

\author{Wonjun Kang\inst{1,3} \and
Geonsu Lee\inst{2,3} \and
Hyung Il Koo\inst{1,4}\and
Nam Ik Cho\inst{3}}

\authorrunning{W. Kang et al.}

\institute{FuriosaAI \and
Neosapience\and
Seoul National University\and
Ajou University}

\maketitle

\begin{figure}[htb!]
\centering
\includegraphics[width=\textwidth]{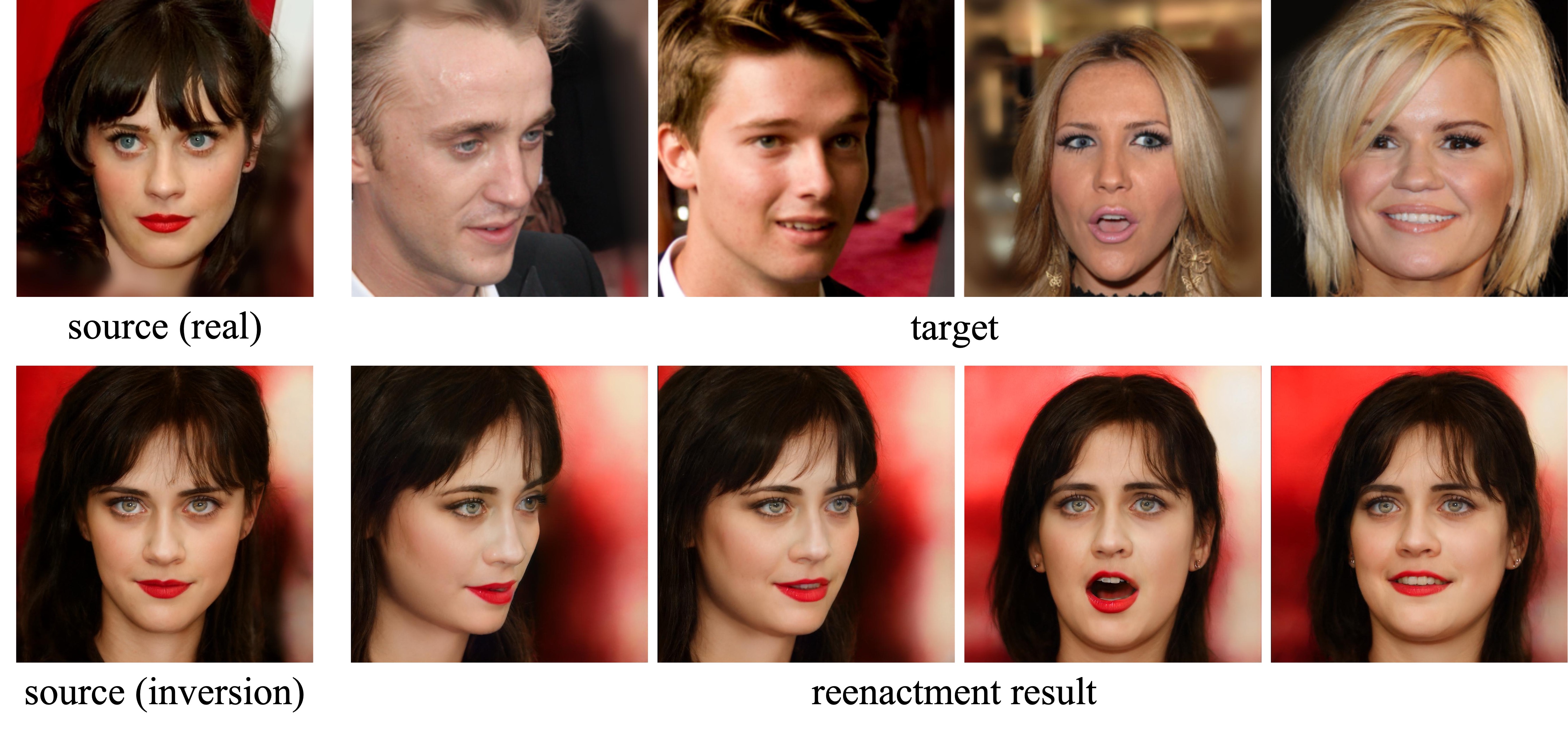}
\caption{Our face reenactment results (1024$\times$1024). The proposed method provides high-resolution results for a variety of poses and expressions.}
\label{fig:examples}
\end{figure}

\begin{abstract}
The goal of face reenactment is to transfer a target expression and head pose to a source face while preserving the source identity. With the popularity of face-related applications, there has been much research on this topic. However, the results of existing methods are still limited to low-resolution and lack photorealism. In this work, we present a  one-shot and high-resolution face reenactment method called MegaFR. 
To be precise, we leverage StyleGAN \cite{karras2019style,karras2020analyzing} by using 3DMM-based rendering images and
overcome the lack of high-quality video datasets by designing a loss function that works without high-quality videos. Also, we apply iterative refinement to deal with extreme poses and/or expressions. Since the proposed method controls source images through 3DMM parameters, we can explicitly manipulate source images. We apply MegaFR to various applications such as face frontalization, eye in-painting, and talking head generation.
 Experimental results show that our method successfully disentangles identity from expression and head pose, and outperforms conventional methods.
\end{abstract}

\section{Introduction}

Face reenactment \cite{thies2016face2face,kim2018deep,wu2018reenactgan,pumarola2018ganimation,zhang2019one,zhang2020freenet,doukas2021headgan,ha2020marionette} is a  face manipulation task that transfers a target facial expression and head pose to a source face while preserving the identity, as demonstrated in Fig. \ref{fig:examples}.
With the ever-increasing interest in virtual humans and the advances in Generative Adversarial Networks (GANs) \cite{goodfellow2014generative},  numerous GAN-based methods have been developed in this field. However, face reenactment results are still limited to low-resolution and lack photorealism.

In the development of a realistic face reenactment system, it is advantageous to exploit a large dataset of a given identity (e.g., a large number of video clips of a single person) \cite{suwajanakorn2017synthesizing,kim2018deep}. However, requirements on such data restrict the applications of the system and recent works focused on face reenactment in a few-shot or one-shot setting \cite{zakharov2019few,burkov2020neural,zakharov2020fast}. However, few-shot face reenactment methods still require videos of (many) people in training and the absence of a large and high-resolution video dataset makes few-shot face reenactment in high resolution a challenging task.

For the synthesis and manipulation of high-resolution face images, we need a well-designed generative model such as StyleGAN \cite{karras2019style,karras2020analyzing} and there have been numerous attempts \cite{nitzan2020face,tewari2020stylerig,deng2020disentangled,tewari2020pie} to disentangle identity from head pose and expression in the latent space for the face reenactment. However, conventional methods still suffer from attribute entangling problems in latent space. To summarize, the main challenges of one-shot and high-resolution face reenactment are (i) the absence of proper large and high-resolution video datasets and (ii) the difficulty in disentangling identity from head pose and expression (or to find meaningful directions in latent space).

In order to alleviate the above challenges, we develop a new method called MegaFR for one-shot and high-resolution face reenactment. Our method exploits StyleGAN and use GAN inversion to perform face reenactment through latent space manipulation. However, unlike existing methods that directly use 3DMM parameters, we use rendering images obtained from 3DMM parameters as network inputs to provide interpretable and explicit information about poses and expressions. Also, we design a loss function that works without high-quality videos.
Finally, we apply an iterative refinement step to deal with extreme head poses and expressions. Our method is designed to control source images with 3DMM parameters, and the proposed method can be considered a controllable StyleGAN as well as a face reenactment method. Experimental results show that our method successfully disentangles identity from expression and head pose, outperforming conventional methods in both visual qualities and control ranges (e.g., extreme poses and/or expressions). Finally, we demonstrate the effectiveness of our work on face-related applications such as face frontalization, eye in-painting, and talking head generation.

\section{Related Work}

For the photo-realistic results, we leverage StyleGAN and latent space manipulation. Therefore, we review latent space manipulation methods and face reenactment in this section.

\subsection{GAN Inversion}

GAN inversion \cite{xia2021gan} aims to map a given real image into the latent space of a pre-trained GAN model. GAN inversion allows us to edit real images by latent space manipulation. In recent years, StyleGAN \cite{karras2019style,karras2020analyzing,karras2020training,karras2021alias} has been commonly used for the pre-trained GAN model because of its state-of-the-art performance and highly disentangled latent space.

To perform GAN inversion, either optimization-based methods \cite{abdal2019image2stylegan,abdal2020image2stylegan++,karras2020analyzing} or learning-based methods \cite{richardson2021encoding,xu2021generative,tov2021designing,alaluf2021restyle} were commonly used: Optimization-based methods search the latent code by minimizing errors directly and learning-based methods train an encoder that yields the latent code of a given input image. GHFeat \cite{xu2021generative} and pSp \cite{richardson2021encoding} used Feature Pyramid Network to extract latent code. In \cite{tov2021designing}, e4e is designed to obtain better editability of reconstructed real images. Also, ReStyle \cite{alaluf2021restyle} introduced an iterative refinement approach to improve reconstruction quality.

\subsection{Latent Space Manipulation}

Recent studies have shown that latent space manipulation techniques are effective in semantic image editing. To discover meaningful directions in latent space, several methods \cite{goetschalckx2019ganalyze,jahanian2019steerability,shen2020interpreting,abdal2021styleflow} were developed in a supervised framework. InterFaceGAN \cite{shen2020interpreting} used attribute classifiers to find the semantic directions in the latent space learned by Support Vector Machine (SVM). StyleFlow \cite{abdal2021styleflow} used Continuous Normalizing Flows (CNF) to find non-linear semantic directions. On the other hand, unsupervised methods \cite{harkonen2020ganspace,shen2021closed} were proposed to find semantic direction without using attribute classifier. Some methods were designed for special purposes such as age transformation \cite{alaluf2021only} or face swap \cite{zhu2021one}. Also, there are methods \cite{xia2021tedigan,patashnik2021styleclip} that focused on text-guided image manipulation for multi-modality. StyleCLIP \cite{patashnik2021styleclip} used CLIP \cite{radford2021learning} loss to perform a text-guided latent space manipulation. Related to our method, recent works \cite{tewari2020stylerig,tewari2020pie} used 3DMM parameters as facial prior to generate and control face images.

\subsection{Face Reenactment}

Face reenactment \cite{thies2016face2face,kim2018deep,wu2018reenactgan,pumarola2018ganimation,zhang2019one,zhang2020freenet,doukas2021headgan,ha2020marionette} is a face manipulation task that transfers a target facial expression and head pose to a source face. Recently, most works focused on one-shot or few-shot face reenactment to alleviate data preparation problems. Several methods \cite{zakharov2019few,burkov2020neural,zakharov2020fast} employed the idea of meta-learning and applied fine-tuning with a few images to perform few-shot face reenactment. PIRenderer \cite{ren2021pirenderer} used 3DMM parameters to enable direct control of head pose and expression. ID-disentanglement \cite{nitzan2020face} was based on StyleGAN and performed one-shot and high-resolution face reenactment.

\section{Method}

Photo-realistic and interpretable manipulation of face images is a challenging task.
We address this problem by employing 3DMM as an intermediate representation, which is a powerful 3D statistical model of human faces \cite{paysan20093d,booth20163d,li2017learning}. 
Fig. \ref{fig:architecture} shows the overview of our MegaFR architecture, where the 3D face reconstruction network yields 3DMM parameters from face images. In this section, we first present our 3D face reconstruction network, which is designed to capture facial expressions faithfully, and explain the overall architecture and the proposed loss function.

\subsection{3D Face Reconstruction Network}
\label{sec:3dmm} 
There are numerous 3D face reconstruction methods based on 3DMM \cite{sanyal2019learning,deng2019accurate,guo2020towards,feng2021learning}, and we can use these off-the-shelf networks for our method. However, in order to perform precise face reenactment, we have trained our 3D face reconstruction network by focusing on capturing facial expressions more accurately.
Among several 3DMM models, we choose FLAME \cite{li2017learning}, which is defined as
\begin{equation}
M(\beta, \psi, \theta):\Re^{|{\beta }| \times |{\psi}| \times |{\theta}|} \to \Re^{N \times 3}
\end{equation}
where ${\beta }\in \Re^{300}$ is an identity vector, ${\psi }\in \Re^{100}$ is an expression vector, ${\theta }\in \Re^{12}$ is a pose vector, and $N=5023$ is the number of vertices. 

Our 3D face reconstruction network is trained with images and landmarks pairs like conventional methods, however, we put more weights on eye and mouth landmarks to capture facial expressions more precisely. For the training, we use a differential renderer to deal with silhouettes of faces in the loss function.

\subsection{Architecture}

\begin{figure}[htb!]
\centering
\includegraphics[width=\textwidth]{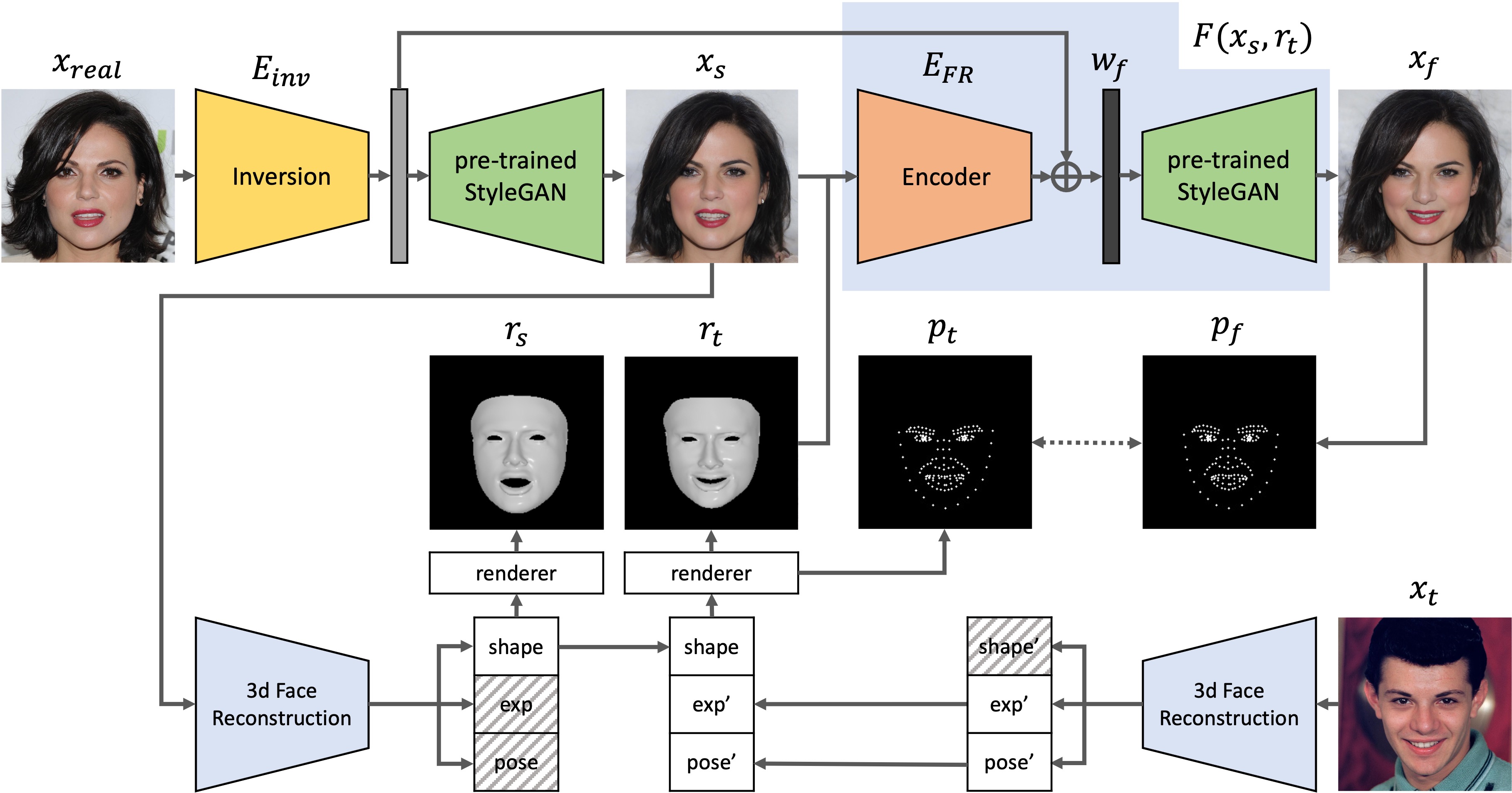}
\caption{Our MegaFR architecture. The network receives a source face image and target 3D rendering image to perform face reenactment.}
\label{fig:architecture} 
\end{figure}

 We use pre-trained StyleGAN to generate high-resolution results and perform face reenactment in the latent space. To this end, we have to map a given image into the latent space of StyleGAN and find its latent code. Among a number of inversion encoders \cite{richardson2021encoding,tov2021designing,alaluf2021restyle,zhu2020domain}, we adopt e4e \cite{tov2021designing} because of its superior editability in latent space.
 
As shown in Fig. \ref{fig:architecture}, our encoder $E_{FR}$ gets a reconstructed source face image $x_{s}$ and a target rendering image $r_{t}$. To generate a target 3D rendering image $r_{t}$, we calculate 3DMM parameters of a reconstructed source face image $x_{s}$ with a pre-trained 3D face reconstruction network (Sec. \ref{sec:3dmm}) 
and extract shape parameters (identity). 
For pose and expression parameters, we apply the reconstruction network to a target image $x_{t}$.
Using these parameters, we obtain a target rendering image $r_{t}$. Then, the reconstructed source face image $x_{s}$ and the target rendering image $r_{t}$ are concatenated and fed into the encoder $E_{FR}$.

We design encoder $E_{FR}$ based on pSp \cite{richardson2021encoding} architecture and modify the number of input channels from 3 to 6. Note that all networks are frozen except the encoder $E_{FR}$ during training. Intuitively, the encoder $E_{FR}$ yields residual latent code, which indicates a direction for expression and head pose changes in latent space. Finally, residual latent code is added to the initial latent code and generates the final image $x_{f}$ with the pre-trained StyleGAN.

As discussed in StyleCLIP \cite{patashnik2021styleclip} and TediGAN \cite{xia2021tedigan}, coarse and medium layers of StyleGAN mainly control head pose and expression, and fine layers are related to colors and other details. Therefore, we partition layers into three groups like StyleCLIP: We neglect fine layers and perform latent manipulation only on coarse and medium layers. This approach helps us control head pose and expression while minimizing the side effects on identity.

\subsection{Loss Function}

Since there are no ground-truth pairs in training our network, we design a loss function that works without ground truth outputs. The loss function is
\begin{equation}
L_{total} (x,r) =\lambda_{1}L_{ID}+\lambda_{2}L_{lmk}+\lambda_{3}L_{pairwise}+\lambda_{4}L_{cyc}+\lambda_{5}L_{self}+\lambda_{6}L_{latent}
\end{equation}
where $x$ and $r$ are an input image and a rendering image respectively.

\subsubsection{ID loss.}
First, we use the ID loss to preserve the identity of the source face,
\begin{equation}
L_{ID}=1-\left \langle R(x_{s}), R( F (x_{s},r_{t})) \right \rangle,
\end{equation}
where $R$ is the pre-trained backbone network of the ArcFace \cite{deng2019arcface}.

\subsubsection{3D facial landmark loss.}
3D facial landmark loss can be used to ensure successful face reenactment from target to source. We obtain 3D facial landmarks using 
the 3D face reconstruction network, and the landmark loss is given by
\begin{equation}
L_{lmk}=\sum_{i}^{}\left \| p_{t}^{i}-p_{f}^{i} \right \|_{1}
\end{equation}
where $p_{t}^{i}$ and $p_{f}^{i}$ are  3D facial landmarks (see Fig. \ref{fig:architecture}) and $i$ is a landmark index.

\subsubsection{Landmark pairwise loss.}
In order to faithfully reflect target expressions, we 
use additional loss for the mouth and eyes. 
Specifically, we calculate the distances between corresponding landmarks along the upper (U) and lower (L) lines of the mouth and eyes respectively:
\begin{equation}
L_{pair}=\sum_{j}^{}\left \| (p_{t}^{j,U}-p_{t}^{j,L})-(p_{f}^{j,U}-p_{f}^{j,L}) \right \|_{1}
\end{equation}
where $p_{t}^{j,U}$, $p_{t}^{j,L}$, $p_{f}^{j,U}$, and $p_{f}^{j,L}$ are corresponding 3D facial landmarks (see Fig. \ref{fig:architecture}).

\subsubsection{Cycle consistency loss.}
In order to address the absence of ground truths, we use a cycle consistency loss, which is commonly used in unpaired image-to-image translation \cite{zhu2017unpaired,kim2017learning,choi2018stargan}:
\begin{equation}
L_{cyc}=L_{rec}(x_{s},F(F(x_{s},r_{t}),r_{s}))
\end{equation}
where $L_{rec} (\cdot,\cdot)$ is the weighted sum of (a)  the pixelwise l2 loss $L_{2}(\cdot,\cdot)$, (b)  the perceptual loss  $L_{LPIPS} (\cdot, \cdot)$ \cite{zhang2018unreasonable}, and (c) $L_{ID} (\cdot,\cdot)$.

\subsubsection{Self-reconstruction loss.}
Additionally, we use a self-reconstruction loss similar to cycle consistency loss
\begin{equation}
L_{self}=L_{rec}(x_{s},F(x_{s},r_{s})).
\end{equation}

\subsubsection{Latent discriminator loss.}
Finally, we use a latent discriminator loss to prevent the latent codes from deviating from the original distribution of latent space:
\begin{equation}
L_{latent}=-D_{latent}(w_{f})
\end{equation}
where $D_{latent}$ is the pre-trained latent discriminator used in e4e \cite{tov2021designing}, and $w_{f}$ is the input latent code of StyleGAN (see Fig. \ref{fig:architecture}).

\subsection{Iterative Refinement}

Inspired by ReStyle \cite{alaluf2021restyle} and HyperStyle \cite{alaluf2021hyperstyle}, we adopt the iterative refinement scheme to improve the reenactment quality. However, 
unlike them, we apply our iterative refinement scheme only to inference.  
That is,  ground-truth output images of face reenactment are not available in our case. Note that ReStyle and HyperStyle aim to GAN inversion and training input images are also ground-truth output images. 

The procedure of iterative refinement can be expressed as
\begin{equation}
x_{s}^{(i+1)} \leftarrow F(x_{s}^{(i)},r_{t})
\end{equation}
where $x_{s}^{(0)} = x_{s}$. Empirically, we find that applying iterative refinement once is sufficient for most cases.

\begin{figure}[htb!]
\centering
\includegraphics[width=\textwidth]{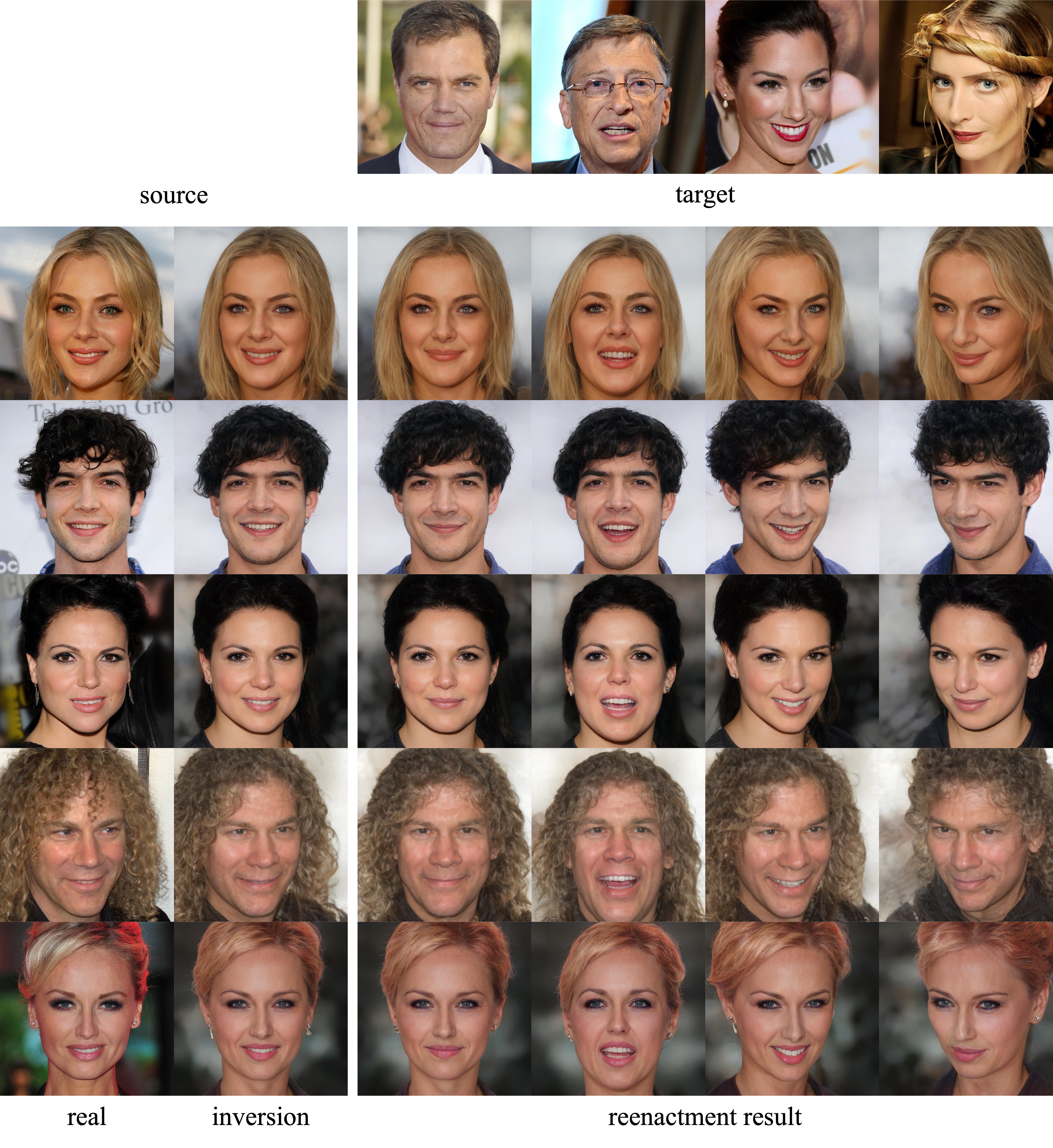}
\caption{Reenactment results on CelebA-HQ. The first and second columns show real ($x_{real}$) and inversion ($x_{s}$) of source faces. The first row shows target faces ($x_{t}$), and other images are results of face reenactment.}
\label{fig:results}
\end{figure}

\section{Experiments}

In experiments, we use FFHQ dataset for training and CelebA-HQ dataset for the test. We use e4e \cite{tov2021designing} as our baseline StyleGAN inversion encoder.
Fig. \ref{fig:results} shows face reenactment results by the proposed MegaFR: Our method successfully reproduces the target's head pose and expression while preserving the source identity and satisfying multi-view consistency.

\begin{figure}[htb!]
\centering
\includegraphics[width=\textwidth]{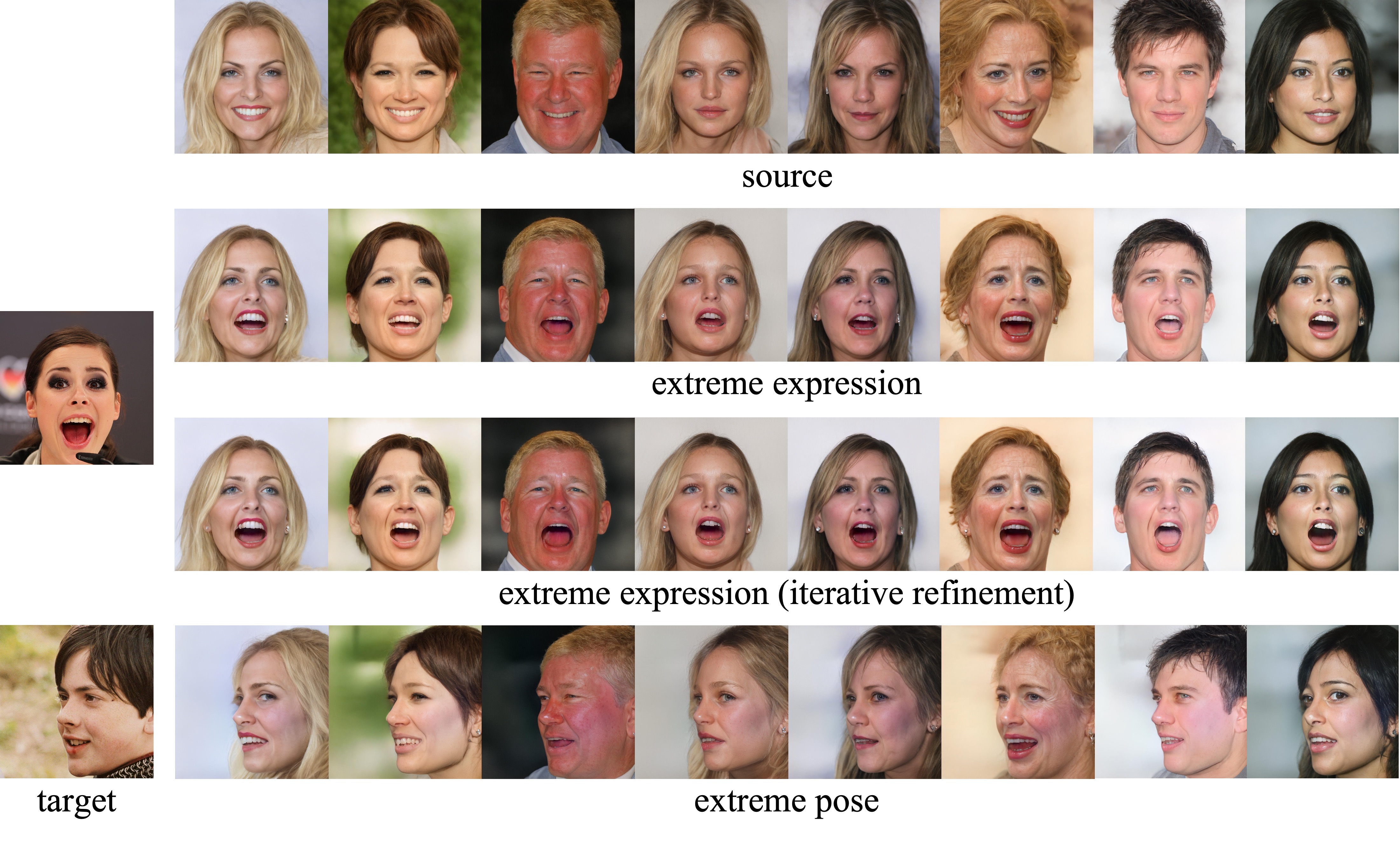}
\caption{Extreme cases and iterative refinement.}
\label{fig:extreme}
\end{figure}

Experimental results on extreme cases are shown in Fig. \ref{fig:extreme}. As shown, our method performs face reenactment successfully even in extreme poses and expressions. As show in the second and third rows, the degrees of mouth opening is not fully reproduced at the first iteration, and it is improved through our iterative refinement.

\subsection{Comparison with Conventional Methods}

\begin{figure}
\captionsetup[subfloat]{farskip=2pt,captionskip=1pt}
\centering
\subfloat[InterFaceGAN \cite{shen2020interpreting} (yaw)]{\label{comparison_interfacegan}\includegraphics[width=\textwidth]{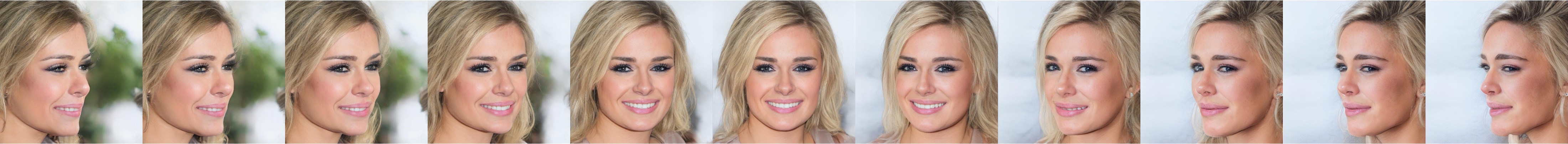}}
\\[-0.01ex]

\subfloat[StyleFlow \cite{abdal2021styleflow} (yaw \& pitch)]{\label{comparison_styleflow}\includegraphics[width=\textwidth]{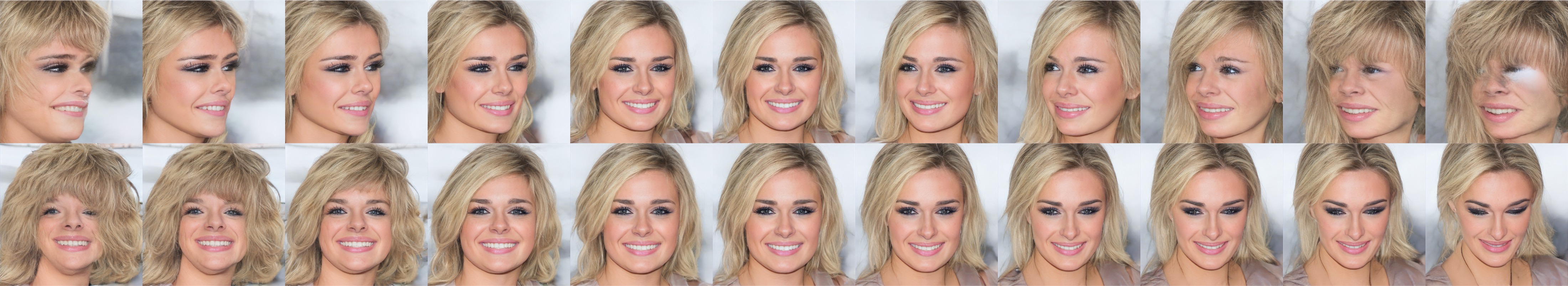}}
\\[-0.01ex]

\subfloat[PIRenderer \cite{ren2021pirenderer} (yaw \& pitch)]{\label{comparison_pirenderer}\includegraphics[width=\textwidth]{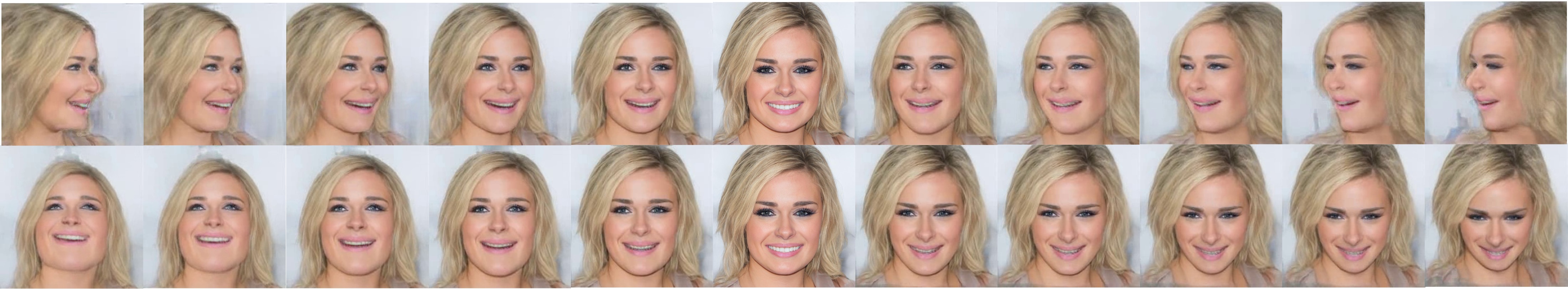}}
\\[-0.01ex]

\subfloat[MegaFR (yaw \& pitch)]{\label{comparison_magafr_1}\includegraphics[width=\textwidth]{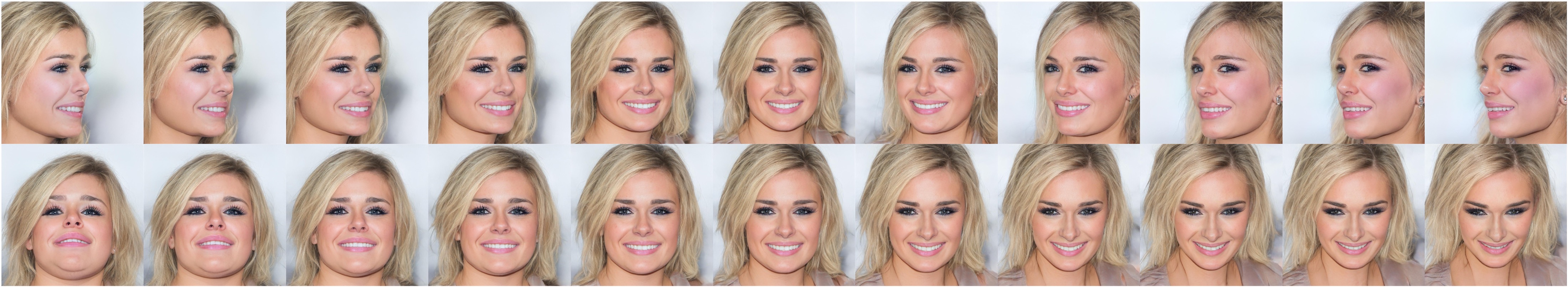}}
\\[-0.01ex]

\subfloat[target for face reenactment]{\label{comparison_target}\includegraphics[width=\textwidth]{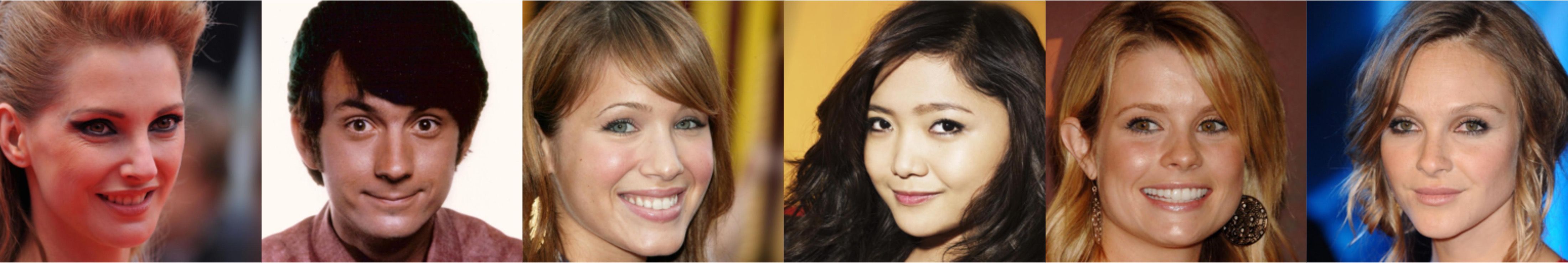}}
\\[-0.01ex]

\subfloat[ID-disentanglement \cite{nitzan2020face} (face reenactment)]{\label{comparison_id-disentanglement}\includegraphics[width=\textwidth]{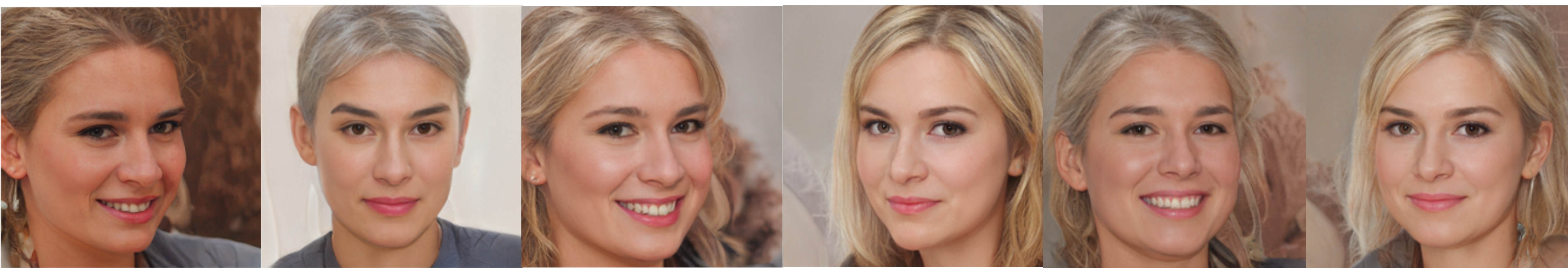}}
\\[-0.01ex]

\subfloat[MegaFR (face reenactment)]{\label{comparison_megafr_2}\includegraphics[width=\textwidth]{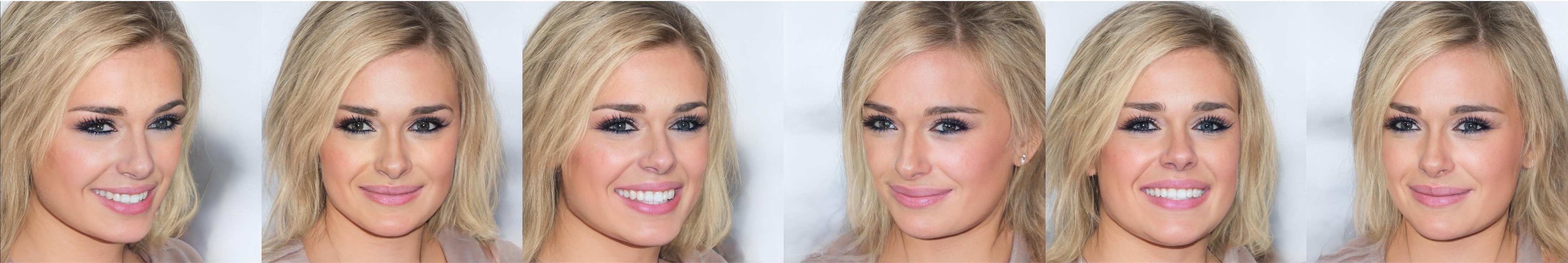}}
\caption{Qualitative comparison. (a)$\sim$(d) show comparison of controlling yaw and pitch. (e)$\sim$(g) show comparison of face reenactment with given various target images.}
\label{fig:comparison}
\end{figure}

Fig. \ref{fig:comparison} shows qualitative comparison between conventional methods and our work: InterFaceGAN \cite{shen2020interpreting}, StyleFlow \cite{abdal2021styleflow}, and PIRenderer \cite{ren2021pirenderer} allow us to control yaw and pitch (InterFaceGAN only provides latent direction of yaw); ID-disentanglement \cite{nitzan2020face} is designed to perform face reenactment with the given source and target images.

InterFaceGAN provides a single latent direction of yaw, which is independent of input images. This approach works on the assumption of linearity of latent space and practically results in entanglement with other attributes such as expression. As shown in Fig. \ref{fig:comparison}(a), InterFaceGAN provides high-resolution results, but the mouth is closing gradually as yaw increases. Although InterFaceGAN also provides a latent direction of a smile, it is not clear to define the degree of smile and we cannot control expression explicitly.

StyleFlow is a non-linear method that uses Continuous Normalizing Flows (CNF) and alleviated the entanglement problem of InterFaceGAN. As shown in Fig. \ref{fig:comparison}(b), the StyleFlow generates high-resolution results and preserves expression for a range of yaw and pitch values. However, StyleFlow only works well in the narrow range of manipulations and fails to handle extreme cases.

Similar to our work, PIRenderer controls the head poses and expressions through 3DMM parameters. However, PIRenderer is trained with low-resolution video datasets and their results are bounded to 256$\times$256. On the other hand, our method is trained with high-resolution image datasets to generate high-resolution results.

ID-disentanglement tried to disentangle identity from head pose and expression, and it can be considered to face reenactment. As shown in Fig \ref{fig:comparison}(d), our method provides better identity disentanglement while ID-disentanglement suffers from identity inconsistency problems.

Table \ref{table:comparison} presents quantitative results of face reenactment methods.
In the table, we evaluate the proposed MegaFR with encoders e4e and pSp, and both encoders outperform the conventional method \cite{nitzan2020face}. 
Although  pSp yields slightly improved metrics, we have found that 
e4e provide visually plausible results and set e4e as a baseline.

\setlength{\tabcolsep}{4pt}
\begin{table}[htb!]
\caption{Quantitative results on CelebA-HQ.}
\begin{center}

\begin{tabular}{|c||c|c|c|}
\hline
Method & FID $\downarrow$ & LPIPS $\downarrow$ & ID similarity $\uparrow$ \\
\hline
ID-disentanglement\cite{nitzan2020face} & 72.7 & 0.46 & 0.33\\
MegaFR (pSp) & \textbf{23.4} & \textbf{0.29} & \textbf{0.47}\\
MegaFR (e4e) & 24.7 & 0.32 & 0.44\\

\hline
\end{tabular}
\end{center}

\label{table:comparison}
\end{table}
\setlength{\tabcolsep}{1.4pt}

\subsection{Editing out-of-domain images}\label{subsec:ood}

Editing faces using reconstructed images ($x_s$ in Fig. \ref{fig:architecture}) as an input to the encoder
sometimes results in slight identity changes or detail loss due to the domain gap between the real  image space and the image space represented by StyleGAN. 

\begin{figure}[htb!]
\centering
\includegraphics[width=\textwidth]{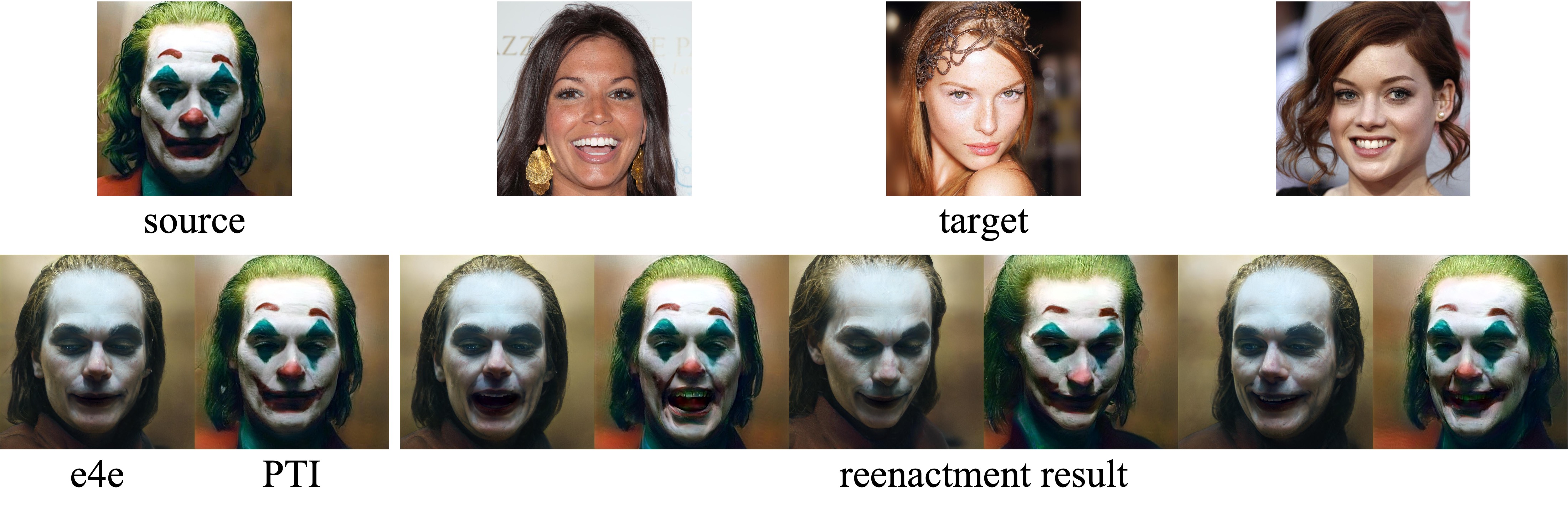}
\caption{Face reenactment results on out-of-domain images.}
\label{fig:ood}
\end{figure}

To bridge this gap, we can apply an additional optimization step such as PTI \cite{roich2021pivotal} to the StyleGAN generator: This approach changes the weight of pre-trained StyleGAN  to preserve a real input image's details. Fig. \ref{fig:ood} shows the face reenactment results on out-of-domain images and we can see that out-of-domain images can be successfully manipulated in our approach.

\section{Applications}

Since our method is designed to control source images through 3DMM parameters, we can control StyleGAN explicitly. In this section, we apply MegaFR to face-related applications and demonstrate the controllability and effectiveness of our work.

\subsection{Face Frontalization}\label{subsec:frontal}

\begin{figure}[htb!]
\captionsetup[subfloat]{farskip=2pt,captionskip=1pt}
\centering
\subfloat[input]{\label{frontalization_input}\includegraphics[width=\textwidth]{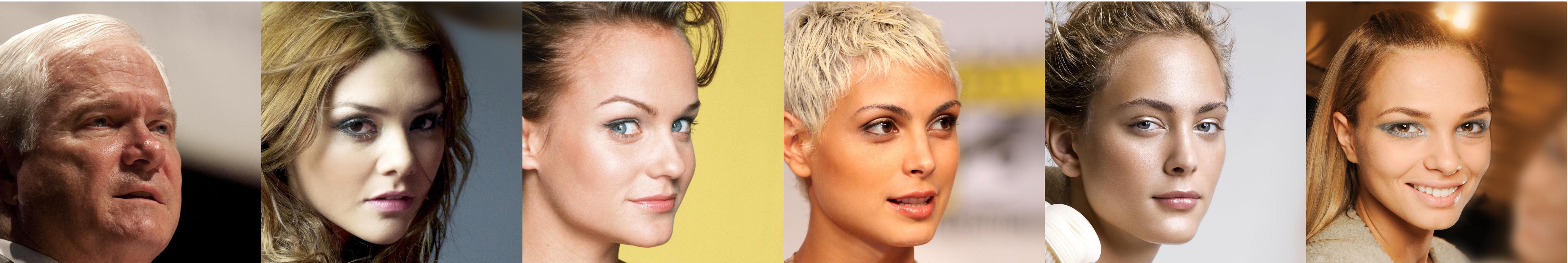}}
\\[-0.01ex]

\subfloat[R\&R \cite{zhou2020rotate}]{\label{frontalization_rnr}\includegraphics[width=\textwidth]{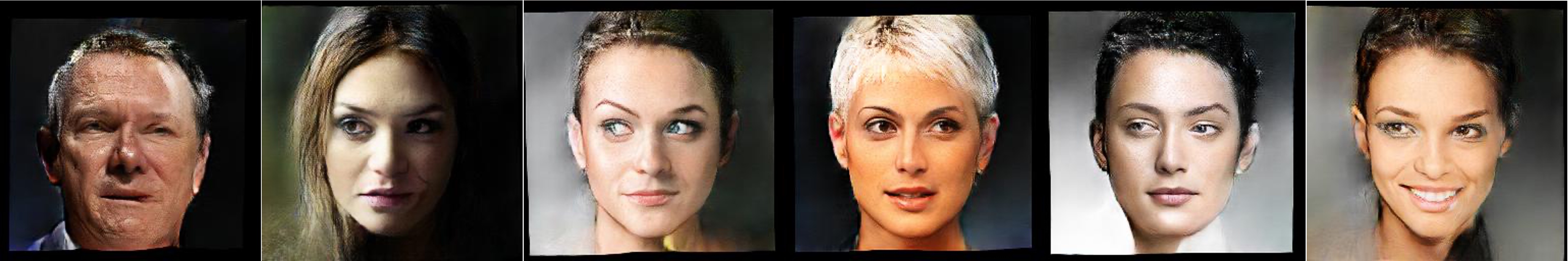}}
\\[-0.01ex]

\subfloat[pSp \cite{richardson2021encoding}]{\label{frontalization_psp}\includegraphics[width=\textwidth]{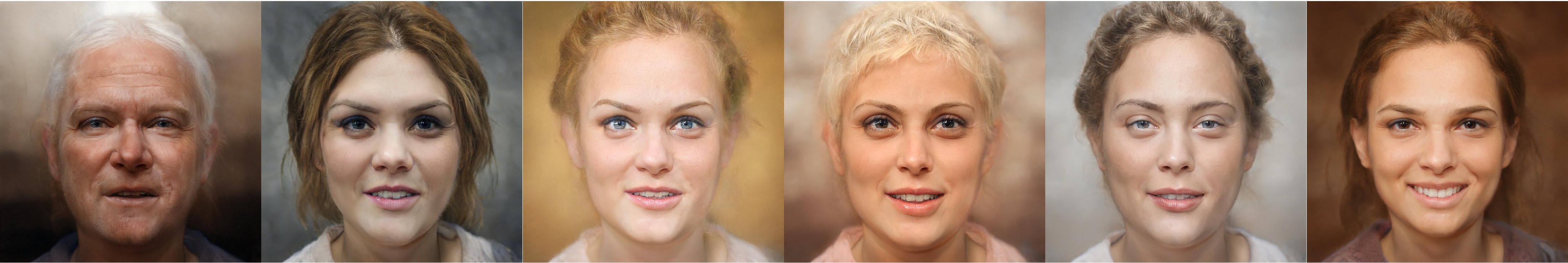}}
\\[-0.01ex]

\subfloat[MegaFR (pSp)]{\label{frontalization_megafr_psp}\includegraphics[width=\textwidth]{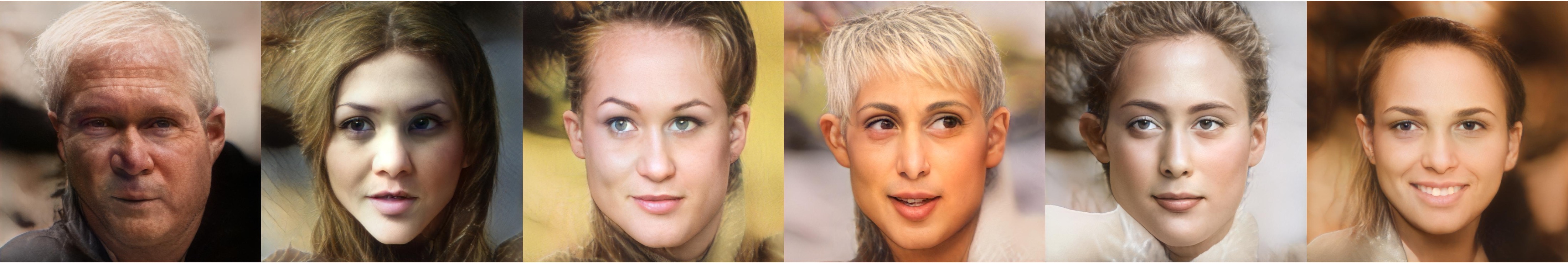}}
\\[-0.01ex]

\subfloat[MegaFR (e4e)]{\label{frontalization_megafr_e4e}\includegraphics[width=\textwidth]{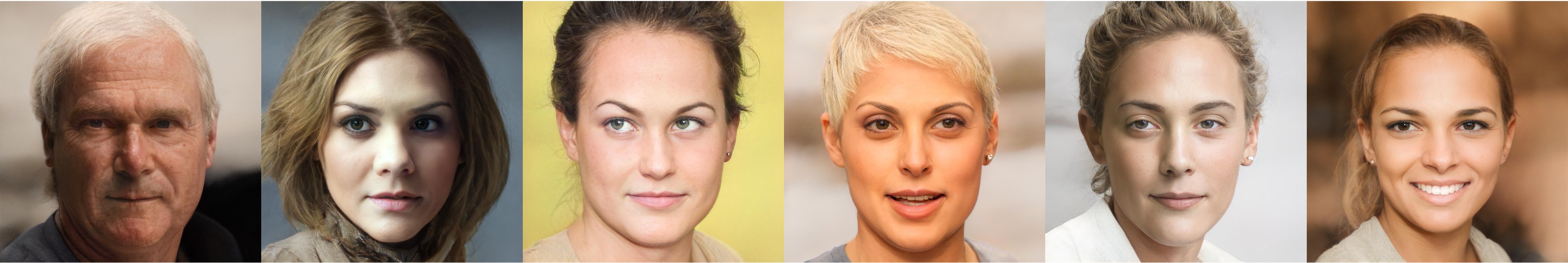}}
\\[-0.01ex]

\subfloat[MegaFR (PTI)]{\label{frontalization_megafr_pti}\includegraphics[width=\textwidth]{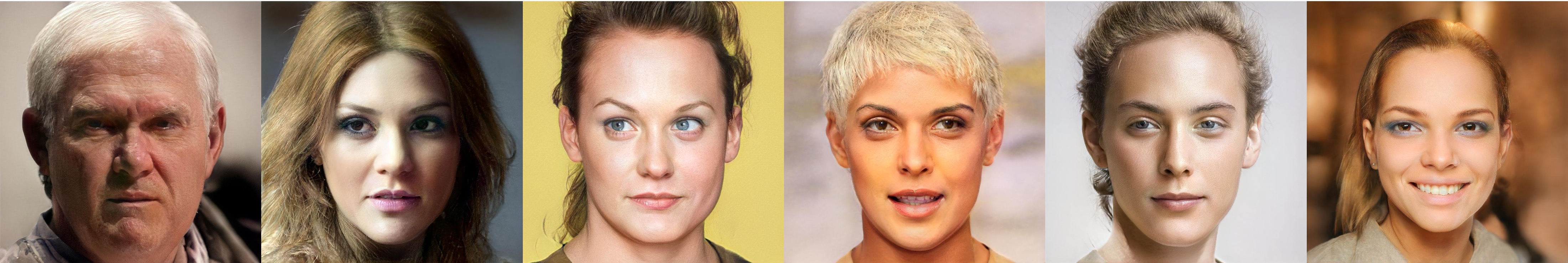}}
\caption{Qualitative comparison of face frontalization on CelebA-HQ.}
\label{fig:frontalization}
\end{figure}

Face frontalization is a face manipulation task that synthesizes a frontal view from an arbitrary posed image. Since paired image datasets are not generally available, 
researchers have attempted to address this problem by leveraging prior knowledge on face images: R\&R \cite{zhou2020rotate} exploited  3DMM and  pSp \cite{richardson2021encoding} exploited the pre-trained StyleGAN.

Compared with conventional methods, our MegaFR allows us to exploit 3DMM and StyleGAN simultaneously. As shown in Fig. \ref{fig:frontalization}, MegaFR satisfies multi-view consistency and visually outperforms other methods. Also, we can reproduce the details of inputs
by employing the additional optimization \cite{roich2021pivotal}.
Table \ref{table:frontalization} shows quantitative comparison of face frontalization methods. As discussed in the previous section, we find that MegaFR (e4e) provides better perceptual quality and editability compared with MegaFR (pSp).

\setlength{\tabcolsep}{4pt}
\begin{table}[htb!]
\caption{Quantitative comparison of face frontalization on CelebA-HQ.}
\begin{center}

\begin{tabular}{|c||c|c|c|}
\hline
Method & FID $\downarrow$ & LPIPS $\downarrow$ & ID similarity $\uparrow$ \\
\hline
R\&R\cite{zhou2020rotate} & 88.0 & 0.49 & \textbf{0.73}\\
pSp\cite{richardson2021encoding} & 62.8 & 0.34 & 0.58\\
MegaFR (pSp) & \textbf{24.3} & \textbf{0.23} & 0.51\\
MegaFR (e4e) & 27.2 & 0.27 & 0.47\\
\hline
\end{tabular}
\end{center}

\label{table:frontalization}
\end{table}
\setlength{\tabcolsep}{1.4pt}

\subsection{Eye In-Painting}

Eye in-painting is a face restoration task that  opens closed eyes in pictures. To this end, most previous works \cite{dolhansky2018eye,yan2020assessing} and commercial applications relied on reference images. Unlike them, we perform eye in-painting without references, since we can control facial expression directly in our framework (through 3DMM parameters).

\begin{figure}[htb!]
\centering
\includegraphics[width=\textwidth]{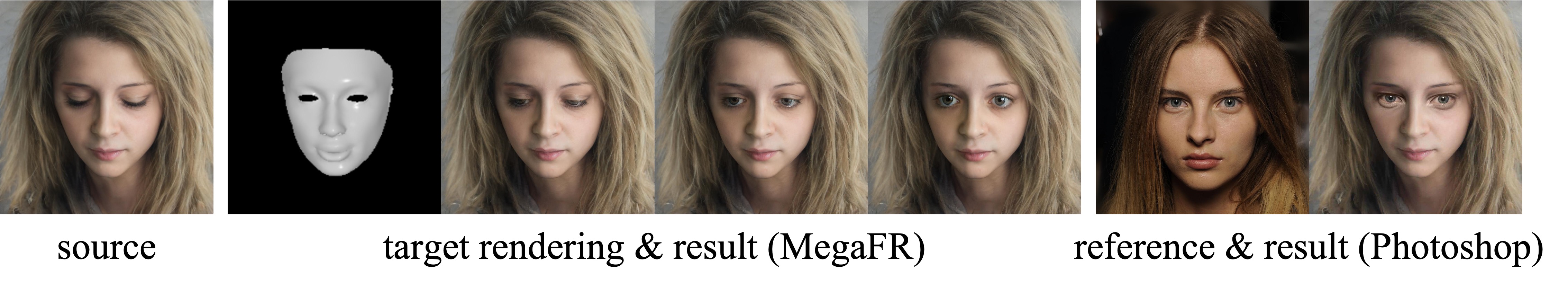}
\caption{Opening closed eyes by MegaFR and Adobe Photoshop.}
\label{fig:eye}
\end{figure}

Fig. \ref{fig:eye} shows the comparison between our method and Adobe Photoshop. As shown, Adobe Photoshop requires additional reference images and yield poor results when there is a discrepancy between source and reference images. On the other hand, our method allows us to open closed eyes without any reference images and control the degree of eye-opening  (by changing 3DMM parameters).

\subsection{Talking Head Generation}

Talking head generation is  a commercially important 
application, and 
many methods have been proposed to provide photo-realistic talking head sequences. 
 Since a huge video set of a single person is not generally available,  recent works \cite{zakharov2019few,burkov2020neural,zakharov2020fast,ren2021pirenderer} have focused on one-shot or few-shot talking head generation. However, to the best of our knowledge, conventional works depend on low-resolution video datasets, and their results are bounded in low-resolution.

\begin{figure}[htb!]
\centering
\includegraphics[width=\textwidth]{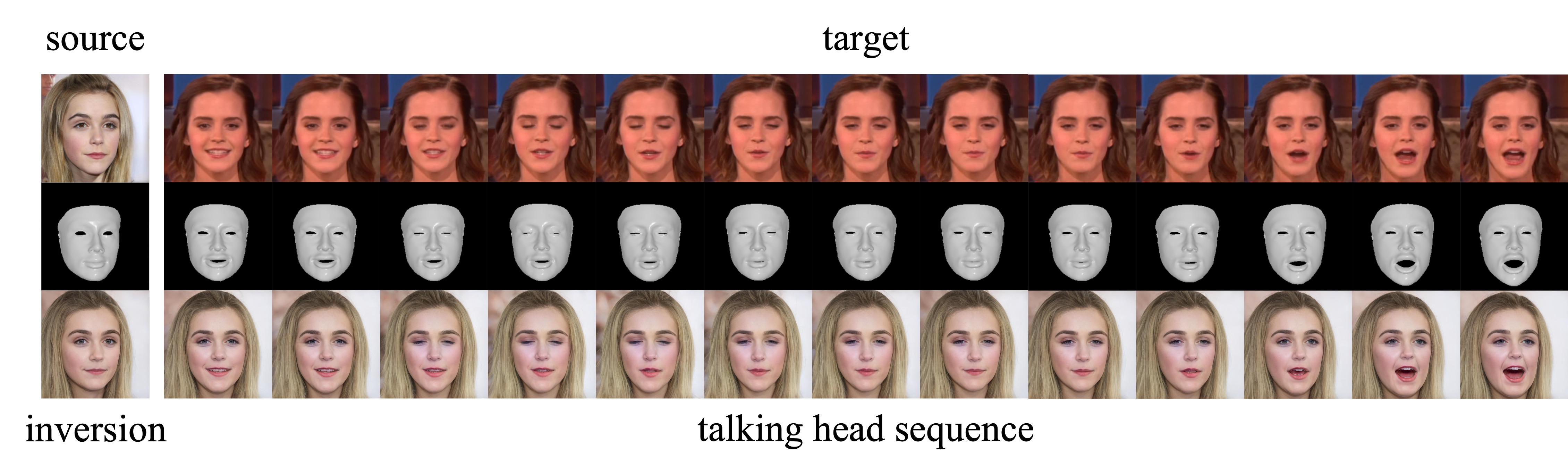}
\caption{Talking head sequence generated by MegaFR.}
\label{fig:talking}
\end{figure}

In this paper, we achieve one-shot and high-resolution face reenactment without video datasets, and we can also apply our method in generating high-resolution talking head sequences. As shown in Fig. \ref{fig:talking}, our method successfully reproduces the target's facial expression and head pose while preserving the source identity. Note that our method not only
reenact  head pose and lip motion but also other facial expression (e.g., eye blinking).

\section{Conclusions}

In this paper, we have proposed a new method for one-shot and high-resolution face reenactment. Instead of using 3DMM parameters as network inputs directly, we used rendering images from 3DMM parameters, which are more CNN-friendly inputs. We designed a loss function to overcome the absence of a high-resolution video dataset and applied an iterative refinement step to deal with extreme cases. Our method successfully disentangles identity from head pose and expression and it enables one-shot face reenactment on megapixels. Furthermore, our method controls source images through 3DMM parameters, and it can 
be used to control StyleGAN explicitly. With the controllability, we have applied MegaFR to face-related applications and demonstrated the effectiveness of our work.

\bibliographystyle{splncs04}
\bibliography{egbib}

\clearpage

\renewcommand{\thesubsection}{\Alph{subsection}.}

\section*{Appendix}

In this appendix, we provide  additional results on CelebA-HQ dataset and FFHQ dataset. All results are in 1024$\times$1024.

\begin{figure}[b!]
\centering
\includegraphics[width=\textwidth]{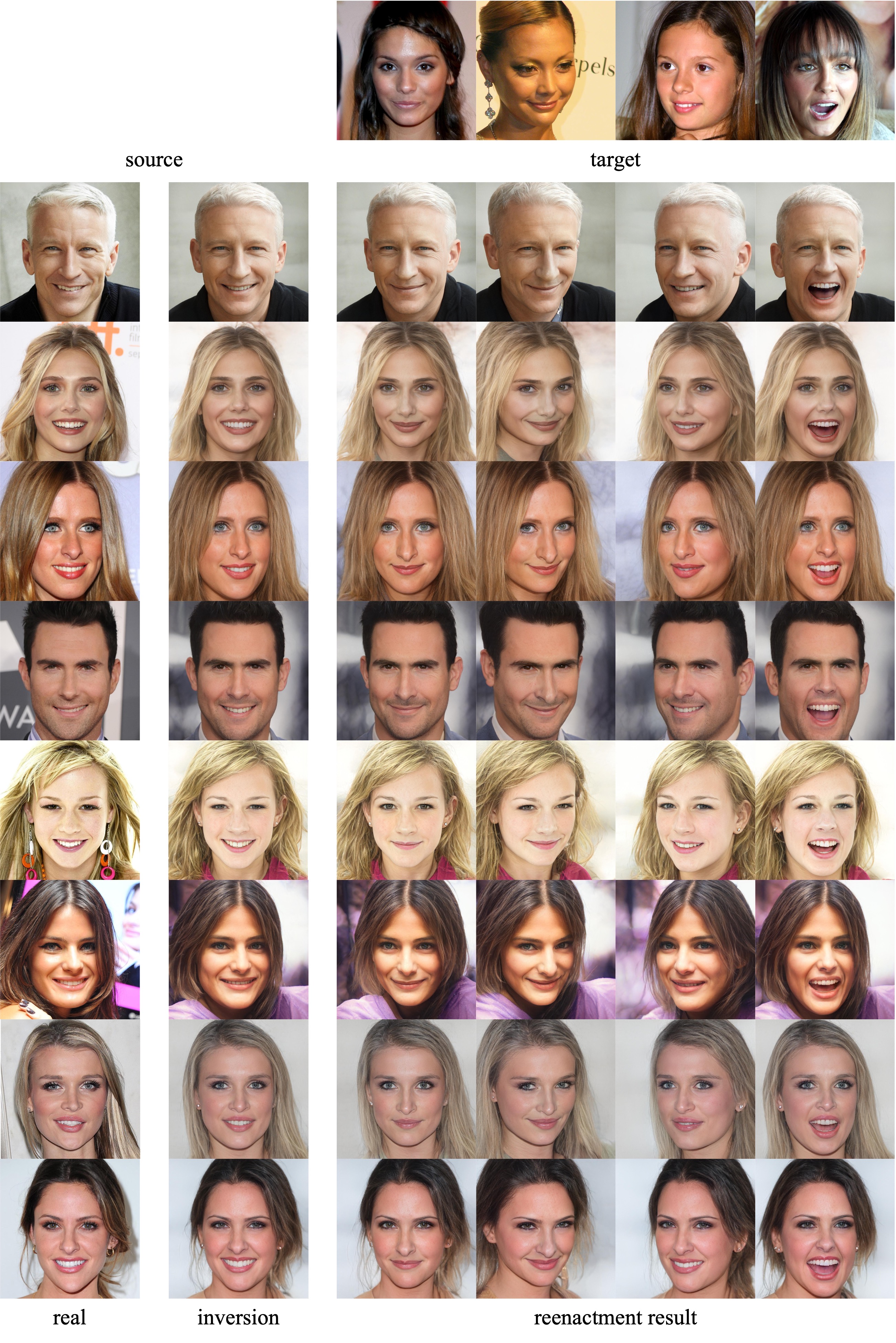}
\caption{Additional face reenactment results (1024$\times$1024) on CelebA-HQ.}
\label{fig:sup_celeb_1}
\end{figure}

\begin{figure}[b!]
\centering
\includegraphics[width=\textwidth]{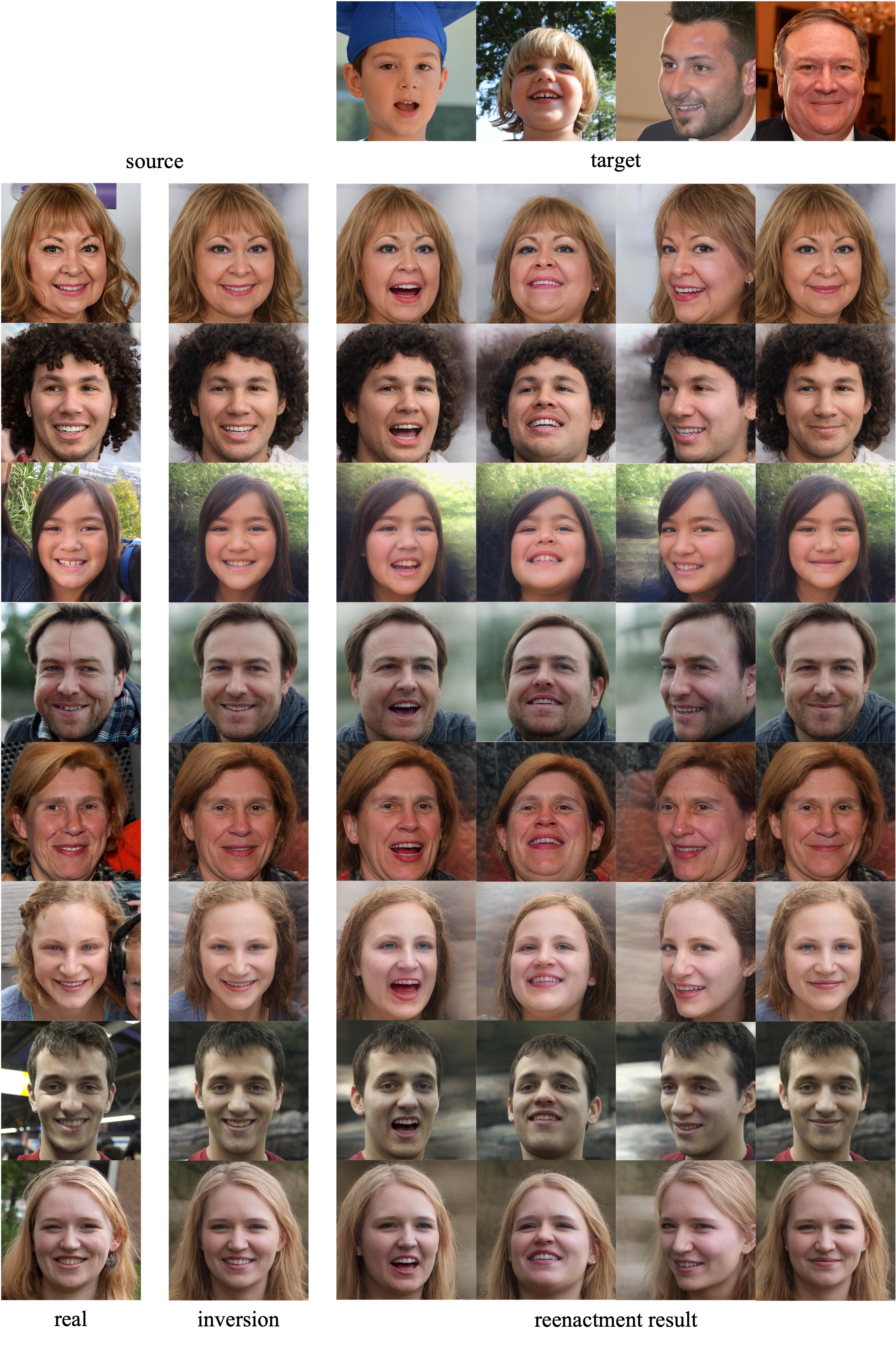}
\caption{Additional face reenactment results (1024$\times$1024) on FFHQ.}
\label{fig:sup_ffhq_1}
\end{figure}

\begin{figure}[b!]
\centering
\includegraphics[width=\textwidth]{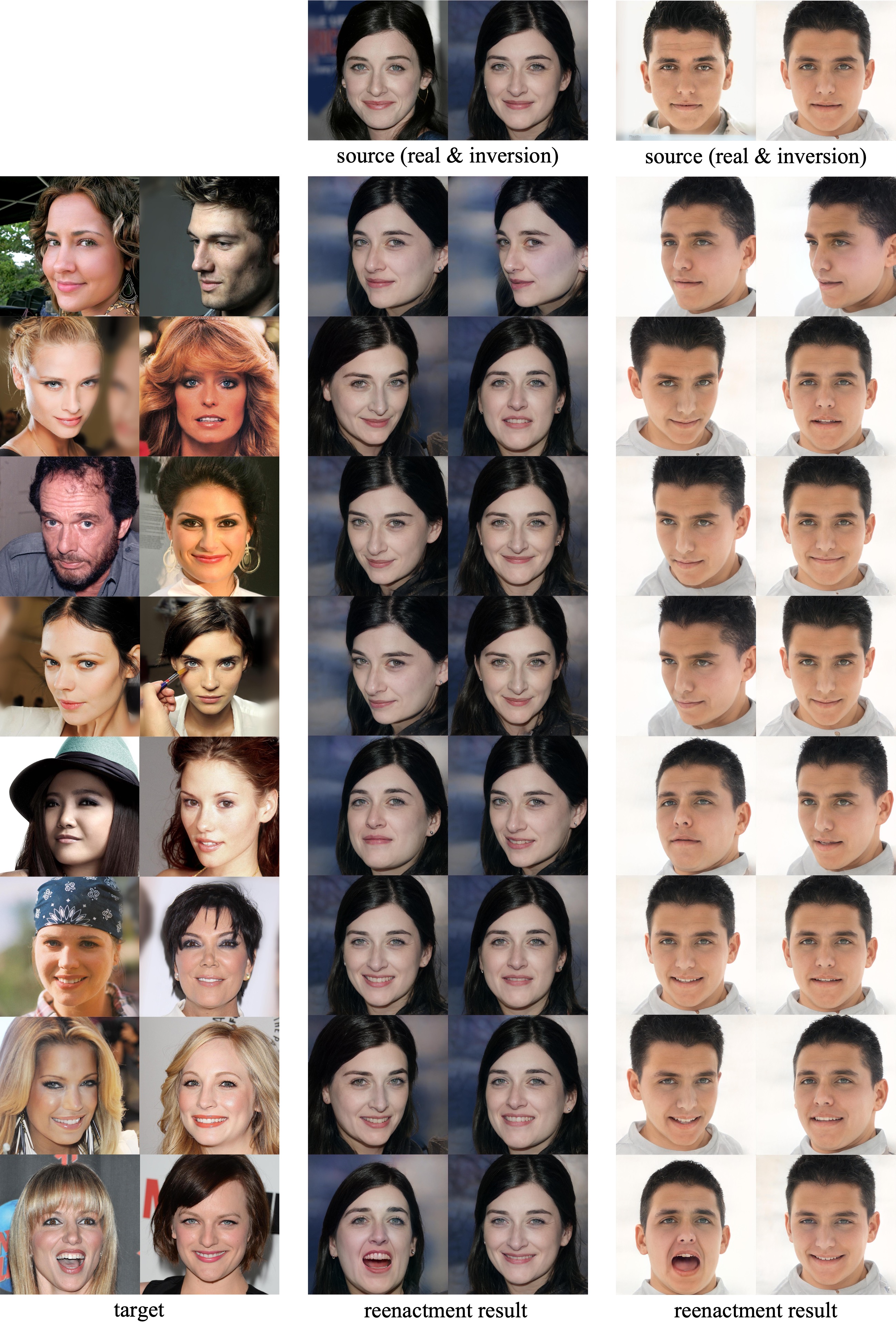}
\caption{Additional face reenactment results (1024$\times$1024) on CelebA-HQ.}
\label{fig:sup_celeb_2}
\end{figure}

\begin{figure}[b!]
\centering
\includegraphics[width=\textwidth]{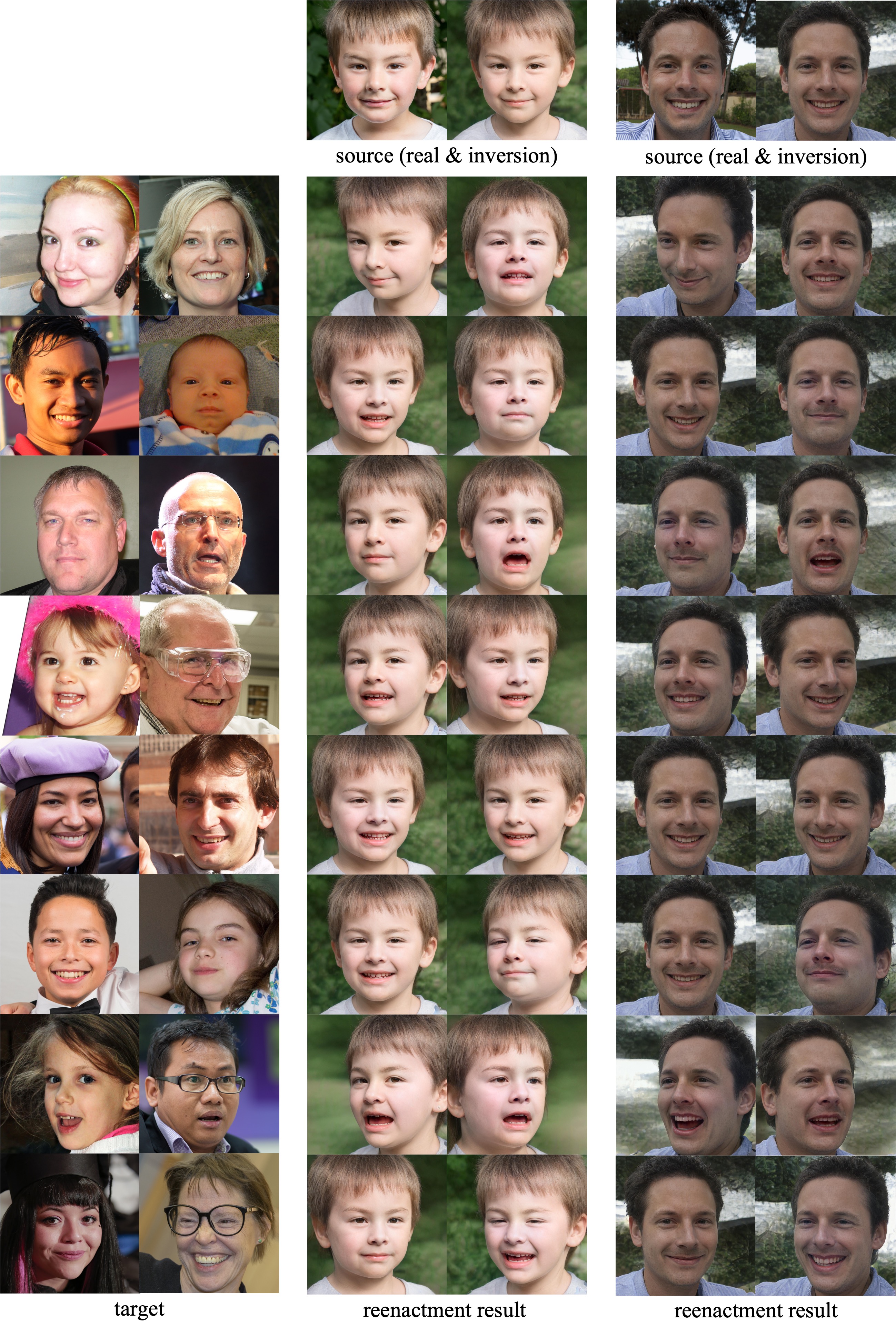}
\caption{Additional face reenactment results (1024$\times$1024) on FFHQ.}
\label{fig:sup_ffhq_2}
\end{figure}

\begin{figure}[b!]
\centering
\includegraphics[width=\textwidth]{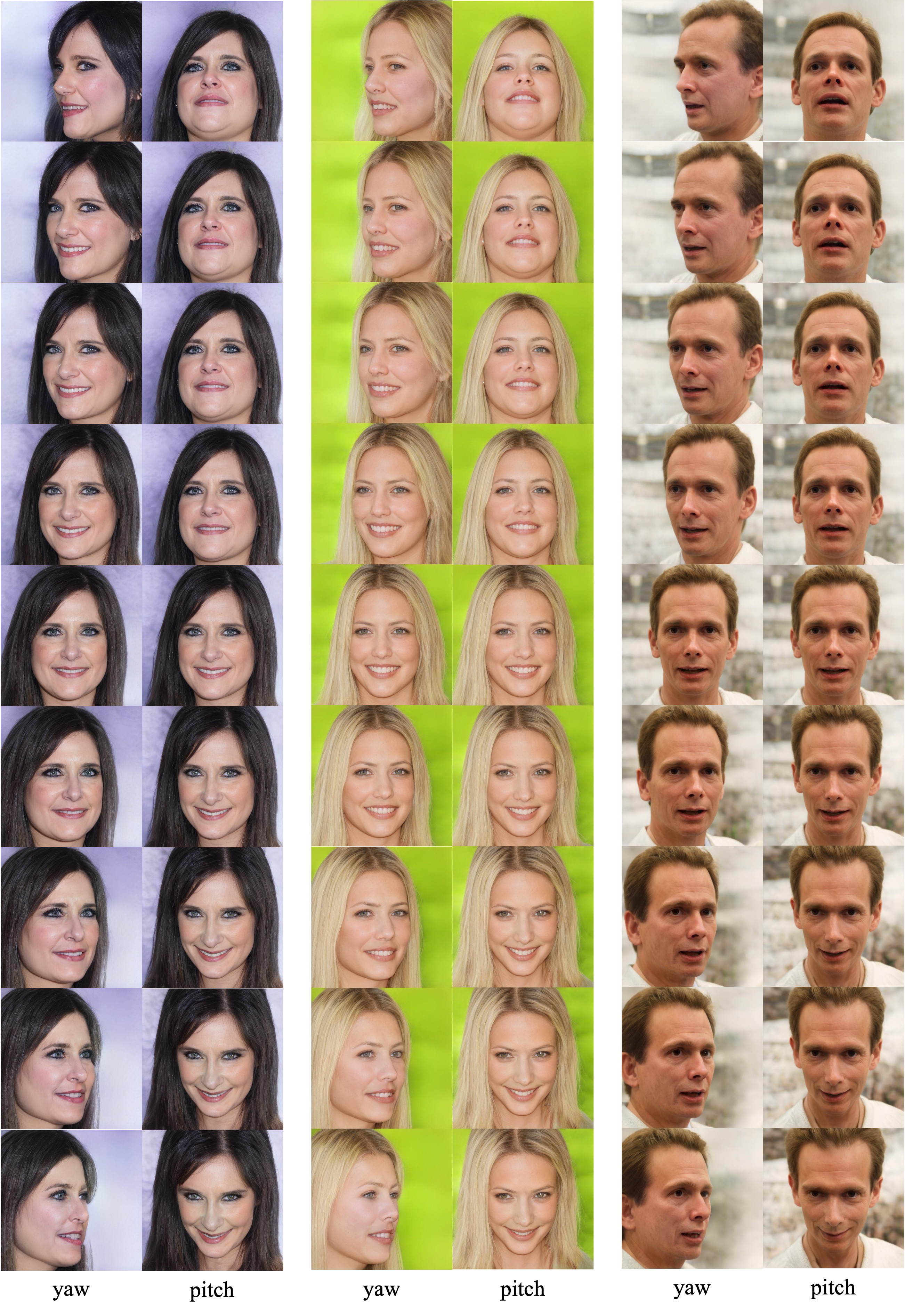}
\caption{Additional controlling yaw \& pitch results (1024$\times$1024) on CelebA-HQ.}
\label{fig:sup_control_celeb}
\end{figure}

\begin{figure}[b!]
\centering
\includegraphics[width=\textwidth]{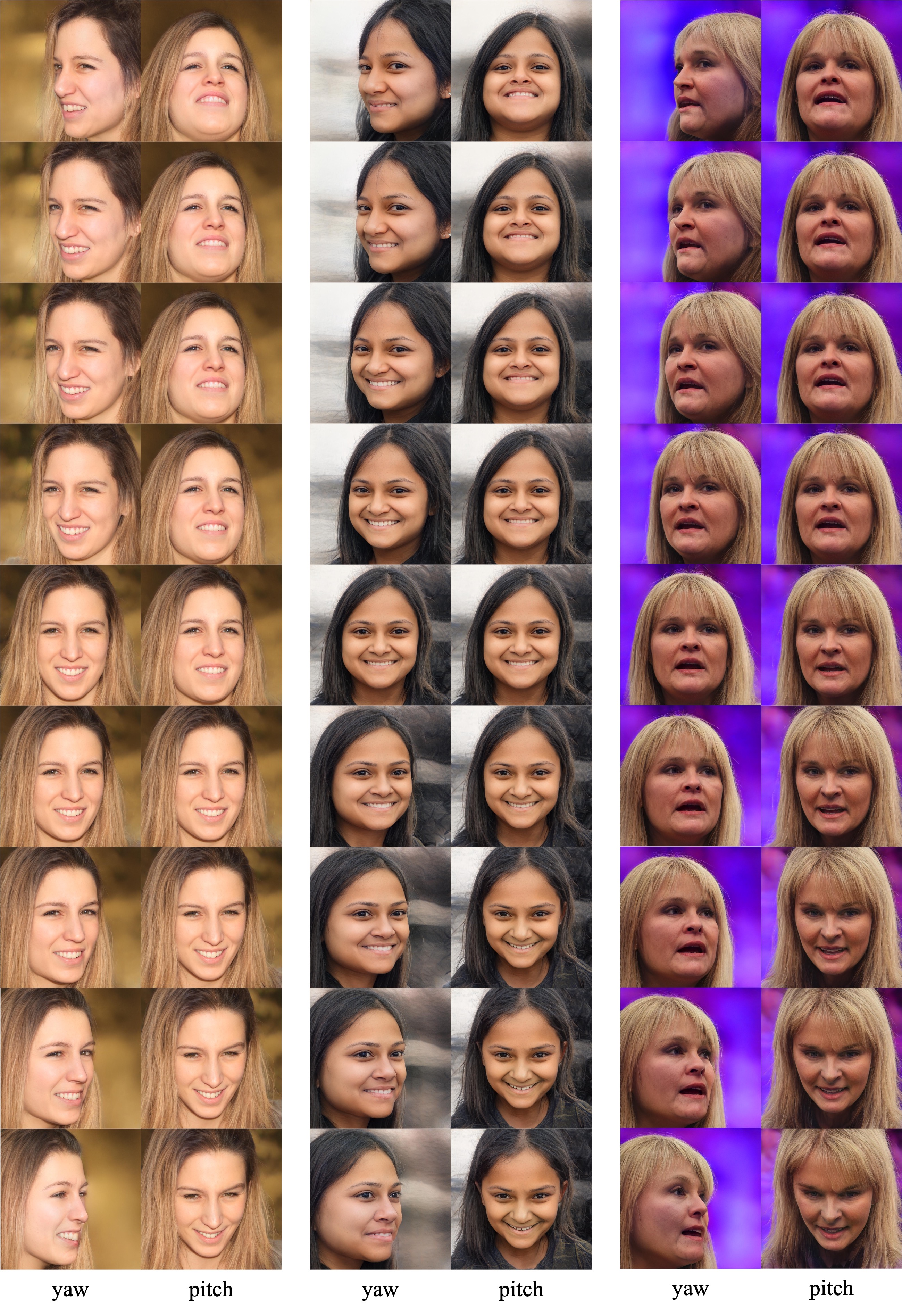}
\caption{Additional controlling yaw \& pitch results (1024$\times$1024) on FFHQ.}
\label{fig:sup_control_ffhq}
\end{figure}

\begin{figure}[b!]
\centering
\includegraphics[width=\textwidth]{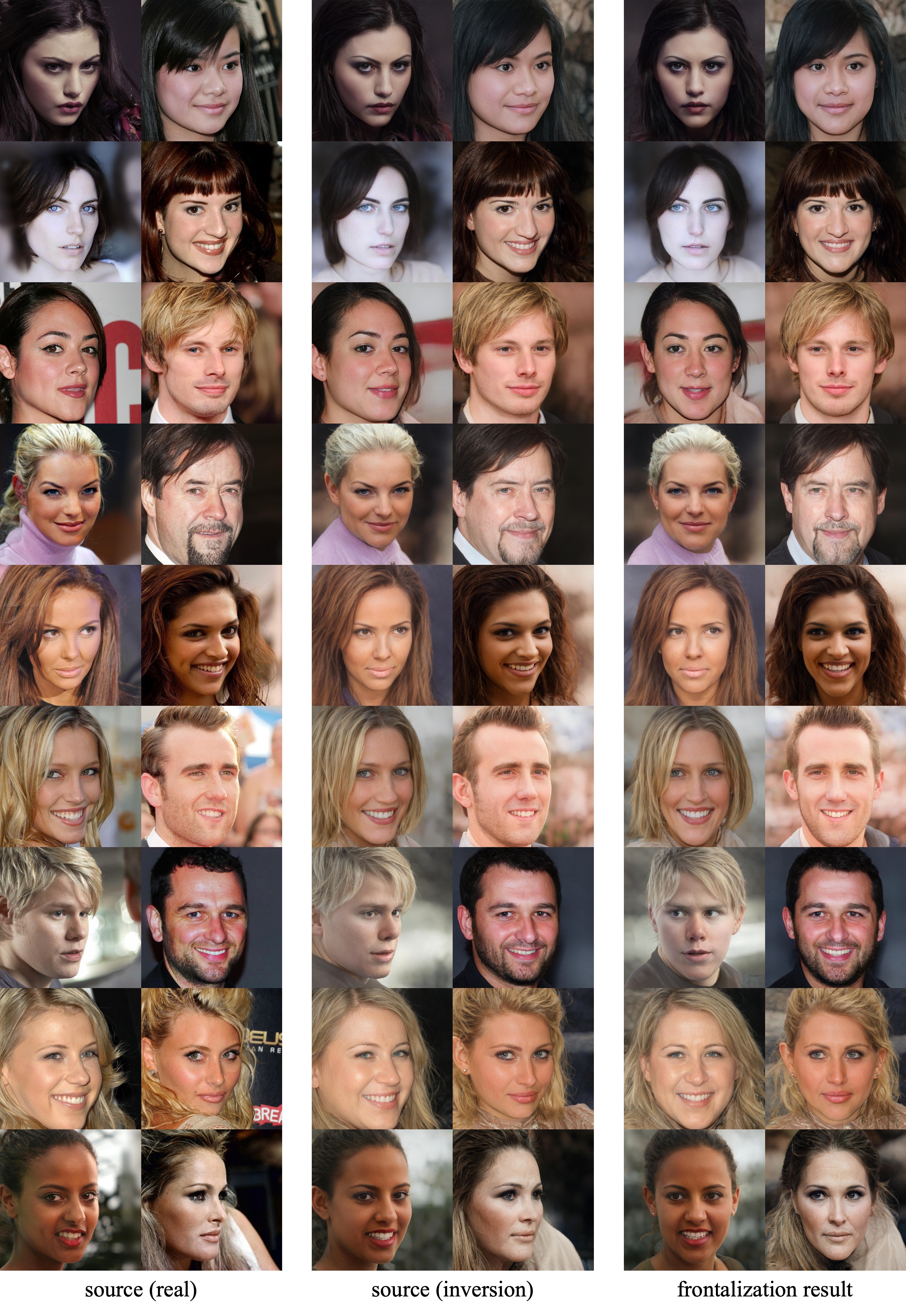}
\caption{Additional face frontalization results (1024$\times$1024) on CelebA-HQ.}
\label{fig:sup_frontal_celeb_1}
\end{figure}

\begin{figure}[b!]
\centering
\includegraphics[width=\textwidth]{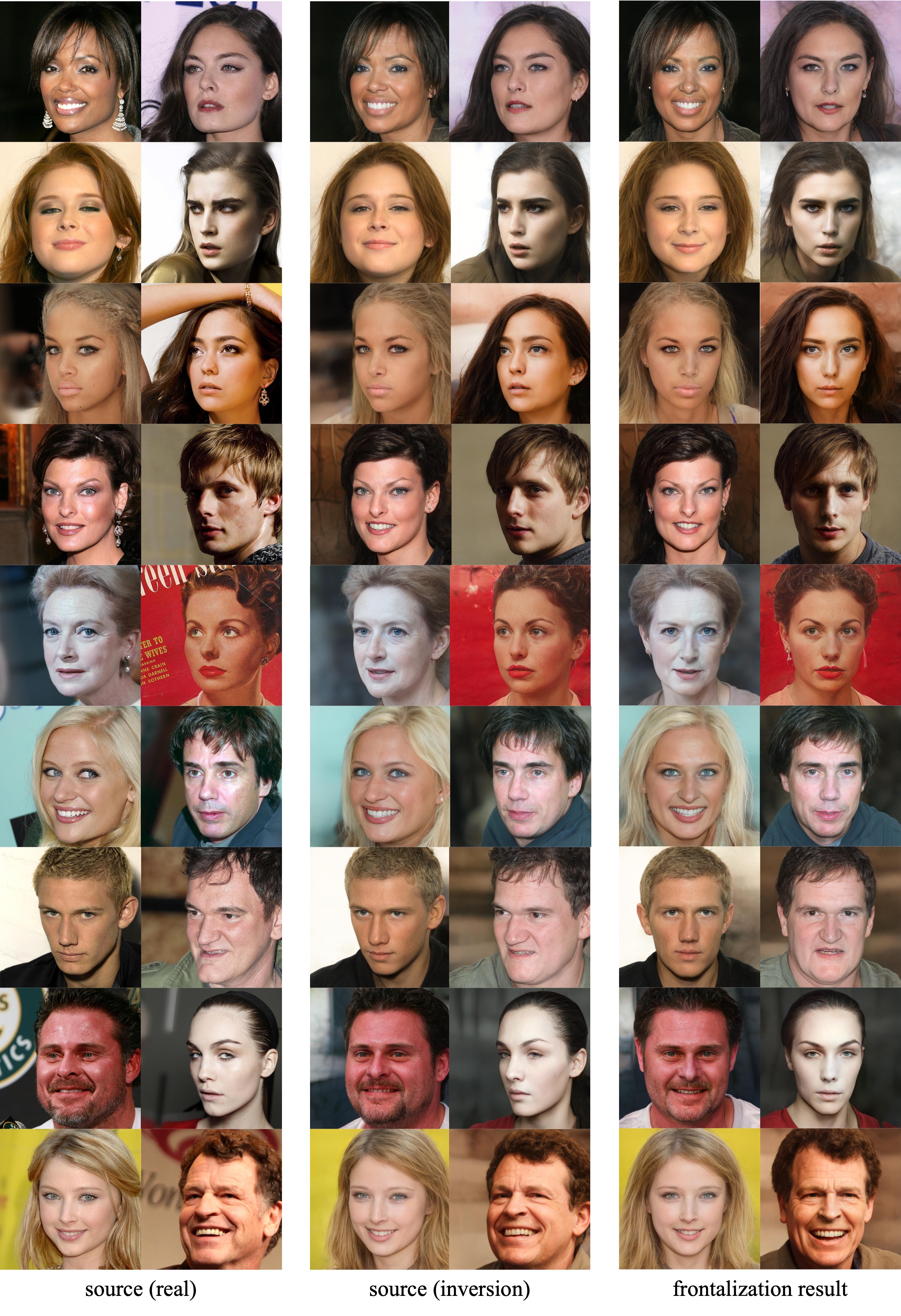}
\caption{Additional face frontalization results on (1024$\times$1024) CelebA-HQ.}
\label{fig:sup_frontal_celeb_2}
\end{figure}

\begin{figure}[b!]
\centering
\includegraphics[width=\textwidth]{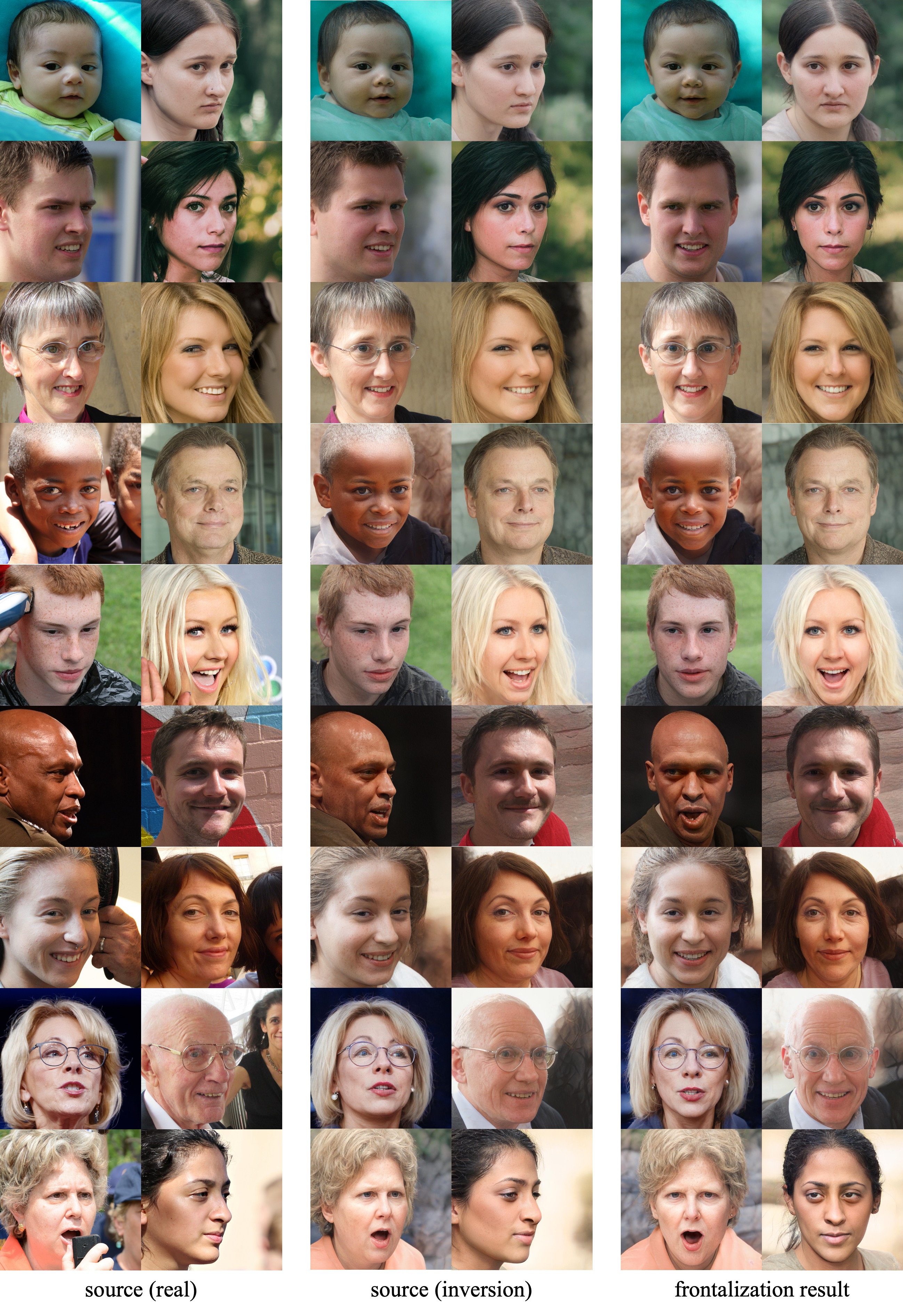}
\caption{Additional face frontalization results (1024$\times$1024) on FFHQ.}
\label{fig:sup_frontal_ffhq_1}
\end{figure}

\begin{figure}[b!]
\centering
\includegraphics[width=\textwidth]{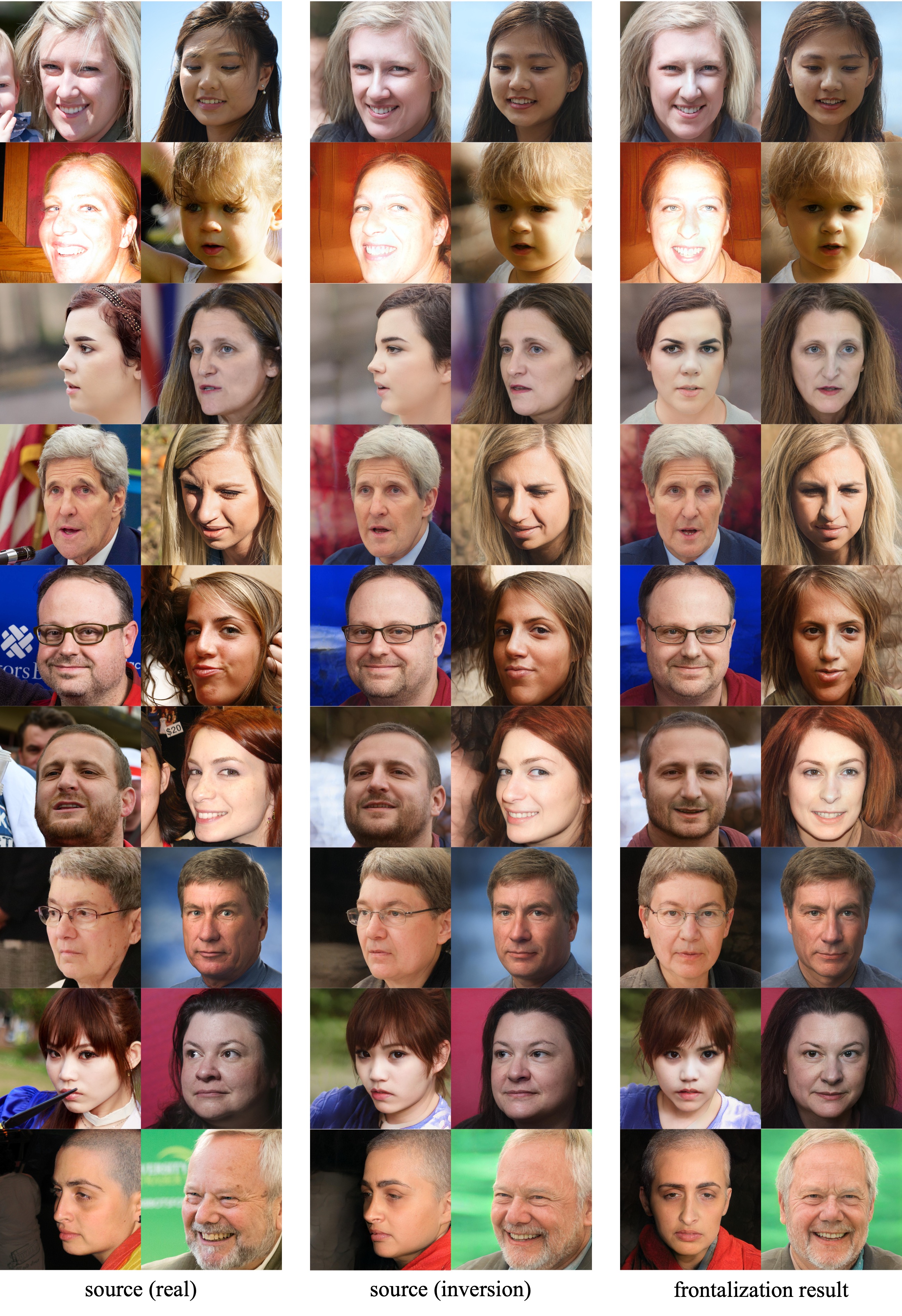}
\caption{Additional face frontalization results (1024$\times$1024) on FFHQ.}
\label{fig:sup_frontal_ffhq_2}
\end{figure}

\begin{enumerate}
    \item Fig. \ref{fig:sup_celeb_1} and Fig. \ref{fig:sup_ffhq_1} show additional face reenactment results on CelebA-HQ and FFHQ. Results contain 8 sources and 4 targets, and they show the effectiveness of MegaFR on a variety of source inputs.
    \\
    \item Fig. \ref{fig:sup_celeb_2} and Fig. \ref{fig:sup_ffhq_2} show additional face reenactment results on CelebA-HQ and FFHQ. Results contain 2 sources and 16 targets, and they show the effectiveness of MegaFR on  a variety of target inputs.
    \\
    \item Fig. \ref{fig:sup_control_celeb} and Fig. \ref{fig:sup_control_ffhq} show additional  yaw and pitch controlling results on CelebA-HQ and FFHQ.
    \\
    \item Fig. \ref{fig:sup_frontal_celeb_1}, \ref{fig:sup_frontal_celeb_2}, \ref{fig:sup_frontal_ffhq_1}, and \ref{fig:sup_frontal_ffhq_2} show additional face frontalization results on CelebA-HQ and FFHQ.
\end{enumerate}

\end{document}